\pdfoutput=1

\documentclass[11pt]{article}

\usepackage{EMNLP2022}

\usepackage{times}
\usepackage{latexsym}

\usepackage[T1]{fontenc}

\usepackage[utf8]{inputenc}

\usepackage{microtype}

\usepackage{booktabs}
\usepackage{microtype}
\usepackage{todonotes}
\usepackage{enumitem}
\usepackage{graphicx}
\usepackage{xspace}
\usepackage{xcolor}
\usepackage{amsmath}
\usepackage{multirow}
\usepackage{multicol}
\usepackage{caption}
\usepackage{hyperref}
\usepackage{subcaption}
\usepackage{colortbl}
\usepackage{framed}
\usepackage{comment}
\usepackage{footmisc}

\newcommand{\firstresult}[1]{\fboxsep0mm\colorbox{gray!30}{\underline{\textbf{#1}}}\xspace}
\newcommand{\secondresult}[1]{\underline{\textbf{#1}}\xspace}
\newcommand{\thirdresult}[1]{\underline{#1}\xspace}

\newcommand{\tanda}{\textsc{TandA}\xspace}
\newcommand{\ie}{\textit{i.e.}\xspace}
\newcommand{\eg}{\textit{e.g.}\xspace}
\newcommand{\ourmodel}{\textsc{Cerberus}\xspace}
\newcommand{\astwo}{AS2\xspace}
\newcommand{\gpd}{IAS2\xspace}
\newcommand{\gpdfull}{Internal Answer Sentence Selection\xspace}

\newcommand{\STAB}[1]{\begin{tabular}{@{}c@{}}#1\end{tabular}}

\newcommand{\albertXXL}{$\textnormal{ALBERT}_\textsc{XXLarge}$\xspace}
\newcommand{\albertB}{$\textnormal{ALBERT}_\textsc{Base}$\xspace}
\newcommand{\robertaL}{$\textnormal{RoBERTa}_\textsc{Large}$\xspace}
\newcommand{\robertaB}{$\textnormal{RoBERTa}_\textsc{Base}$\xspace}
\newcommand{\electraL}{$\textnormal{ELECTRA}_\textsc{Large}$\xspace}
\newcommand{\electraB}{$\textnormal{ELECTRA}_\textsc{Base}$\xspace}

\definecolor{magenta}{HTML}{DC267F}
\definecolor{yellow}{HTML}{FFB000}
\definecolor{cyan}{HTML}{648FFF}

\title{Ensemble Transformer for Efficient and Accurate Ranking Tasks: \\an Application to Question Answering Systems}

\author{Yoshitomo Matsubara~\thanks{~~This work was done while the author was an intern at Amazon Alexa AI.} \\
  University of California, Irvine \\
  \texttt{yoshitom@uci.edu} \\\And
  Luca Soldaini \\
  Allen Institute for AI \\
  \texttt{lucas@allenai.org} \\\AND
  Eric Lind \\
  Amazon Alexa AI \\
  \texttt{ericlind@amazon.com} \\\And
  Alessandro Moschitti \\
  Amazon Alexa AI \\
  \texttt{amosch@amazon.com} \\
}

\begin{document}
\maketitle
\begin{abstract}
Large transformer models can highly improve Answer Sentence Selection (\astwo) tasks, but their high computational costs prevent their use in many real-world applications. 
In this paper, we explore the following research question: \textit{How can we make the \astwo models more accurate without significantly increasing their model complexity?}
To address the question, we propose a Multiple Heads Student architecture (named \ourmodel), an efficient neural network designed to distill an ensemble of large transformers into a single smaller model.
\ourmodel consists of two components: a stack of transformer layers that is used to encode inputs, and a set of ranking heads; unlike traditional distillation technique, each of them is trained by distilling a different large transformer architecture in a way that preserves the diversity of the ensemble members.
The resulting model captures the knowledge of heterogeneous transformer models by using just a few extra parameters.
We show the effectiveness of \ourmodel on three English datasets for \astwo;
our proposed approach outperforms all single-model distillations we consider, rivaling the state-of-the-art large \astwo models that have $2.7\times$ more parameters and run $2.5\times$ slower.
Code for our model is available at \href{https://github.com/amazon-research/wqa-cerberus}{\url{https://github.com/amazon-research/wqa-cerberus}}.
\end{abstract}

\section{Introduction}
\label{sec:introduction}

Answer Sentence Selection (\astwo) is a core task for designing efficient retrieval-based Web QA systems: 
given a question and a set of answer sentence candidates (\eg, retrieved by a search engine), \astwo models select the sentence that correctly answers the question with the highest probability. 
\astwo research originated from the TREC competitions~\cite{wang2007jeopardy}, which targeted large amounts of unstructured text.
\astwo models are very efficient, and can enable Web-powered question answering systems of real-world virtual assistants such as Alexa, Google Home, Siri, and others.

As most research areas in text processing and retrieval, \astwo has been dominated by the use of ever larger transformer model architectures~\cite{vaswani2017attention}.
These models are typically pre-trained using language modeling tasks on large amounts of text~\cite{devlin2019bert,liu2019roberta,conneau2019unsupervised},
and then fine-tuned on specific downstream tasks~\cite{glue,wang2019superglue,hu2020xtreme}.
\citet{garg2020tanda} achieved an impressive accuracy by fine-tuning pre-trained Transformers to the \astwo task on the target datasets.
They established the new state of the art performance for \astwo using a \robertaL model.

\begin{figure}[t]
    \centering
    \includegraphics[width=0.7\linewidth]{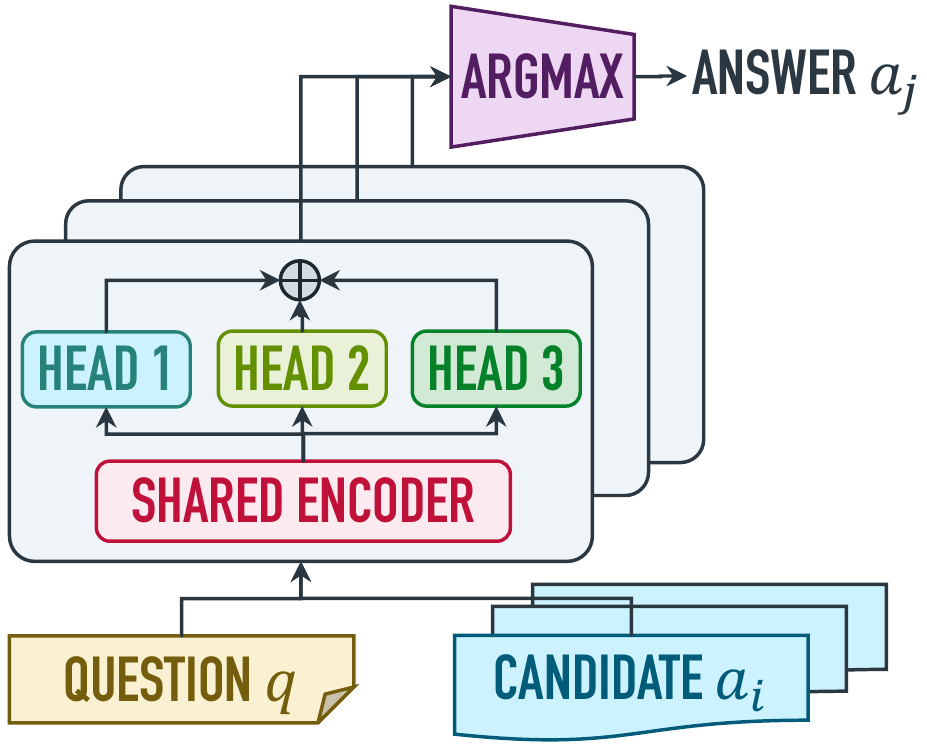}
    \caption{\ourmodel model for answer sentence selection. The model consists of a shared encoder body and multiple ranking heads. \ourmodel independently scores up to hundreds candidate answers $a_i$ for question $q$; The one with highest likelihood is selected as answer.}
    \label{fig:our_model_overview}
\end{figure}

Unfortunately, larger transformer models come at a cost:
they require large computing resources, consume a lot of energy (critically impacting the environment~\cite{strubell2019energy}), and may have unacceptable latency and/or memory usage.
These downsides are critical for \astwo applications, where, for any given query, a model is required to score hundreds or thousands of candidates to select the top-$k$ answers. 
Therefore, in this work, we investigate how \astwo models can be made more accurate without significantly increasing their complexity.

Previous work has addressed the general problem of high computational cost of transformer models by developing techniques for reducing their overall size while maintaining most of their performance ~\citep{polino2018model,liu2018rethinking,li2020train}.
In particular, Knowledge Distillation (KD) techniques have been shown to be particularly effective~\citep{sanh2019distilbert,turc2019well,sun2019patient,sun2020mobilebert,yang2020model,jiao2020tinybert}.
KD techniques use a larger model, known as a \textit{teacher}, to obtain a smaller and thus more efficient model, known as a \textit{student} \cite{hinton2015distilling}.
The student is trained to mimic the output of the teacher.
However, we empirically show that, at least for \astwo, \textsc{Base} models trained through distillation are still significantly behind the state of the art, \ie, models based on \textsc{Large} transformers.

In this paper, we introduce a new transformer model for \astwo that matches the state of the art while being dramatically more efficient.
Our main idea is based on the following considerations: 
first, in recent years, several transformer model families have been introduced, each pretrained using different datasets and modeling techniques~\citep{rogers2021primer}. 
Second, ensembling several diverse models has shown to be an effective way to improve performance in many question answering and ranking tasks \citep{xu2020improving,zhang2020query,Liu2020RikiNetRW,Lin2020TradeoffsIS}.
Our contribution lies in a new approach to approximate a computationally expensive ranking ensemble into a single efficient architecture for \astwo tasks. 

More specifically, our investigation proceeds as follows.
First, we optimize ranking architectures for \astwo by training $k$ student models to replicate $k$ unique teacher architectures.
When ensembled, we show that they achieve better performance than any standalone models at the cost of increased computational burden.
Then, to preserve the accuracy of this ensemble while achieving lower complexity, we propose a new \textit{Multiple Heads Student} architecture, which we refer to as \ourmodel.
As shown in Fig.~\ref{fig:our_model_overview}, \ourmodel is composed of a shared encoder body and multiple ranking heads.
The encoder body is designed to derive a shared representation of input sequences, which gets fed to ranking heads.
We show that if each ranking head is trained to mimic a unique teacher distribution, it is possible to achieve the desirable diversity through ensemble model while being significantly more efficient.

We train a \ourmodel model using three different teachers: RoBERTa~\cite{liu2019roberta}, ELECTRA~\cite{clark2019electra}, and ALBERT~\cite{lan2019albert}.
We conduct experiments on three \astwo datasets: ASNQ~\cite{garg2020tanda}, WikiQA~\cite{yang2015wikiqa}, and an internal corpus (\gpd).
Our results show
that \ourmodel consistently improves over all models trained with single teachers, rivaling performance of much larger models including multiple variants of ensemble models; further, \ourmodel matches current state-of-the-art \astwo models (\tanda by~\citet{garg2020tanda}), while saving 64\% and 60\% in model size and latency, respectively.

In summary, our contribution is four-fold:
\begin{enumerate}[topsep=0pt, partopsep=1em, leftmargin=2em, label=(\roman*), itemsep=-0.1em]
    \item We propose \ourmodel, an efficient architecture specifically designed to distill an ensemble of heterogeneous transformer models into a single transformer model for \astwo tasks while preserving ensemble diversity.
    \item We conduct large-scale experiments with multiple transformer model families and show that \ourmodel improves performance of equally sized distilled model, rivaling much larger ensemble and state-of-the-art \astwo models.
    \item We discuss various training methods for \ourmodel and show three key factors to improve \astwo performance: (\textit{a}) multiple ranking heads in \ourmodel, (\textit{b}) multiple teachers, and (\textit{c}) heterogeneity in teacher models.
    \item We present a comprehensive analysis of the \ourmodel, both in terms of ranking behavior and efficiency, highlighting the effect of several design decisions on its performance.
\end{enumerate}

\section{Related Work}
\subsection{Answer Sentence Selection (\astwo)}
Several approaches for \astwo have been proposed in recent years.
\citet{severyn2015learning} used CNNs to learn and score question and answer representations, while others proposed alignment networks \citep{shen-etal-2017-inter,tran-etal-2018-context,DBLP:journals/corr/abs-1806-00778}.
Compare-and-aggregate architectures have also been 
extensively studied \citep{wang2016compare,Bian:2017:CMD:3132847.3133089,DBLP:journals/corr/abs-1905-12897,matsubara2020reranking}. 
\citet{madabushi2018integrating} exploited fine-grained question classification to further improve answer selection.
\citet{garg2020tanda} have achieved impressive performance by fine-tuning transformer models using a novel transfer-and-adapt technique. 
\citet{lauriola2021answer} and \citet{han2021modeling} leveraged contextual information as an additional input to improve model accuracy for \astwo tasks.

\subsection{Single Model Distillation}
Knowledge distillation for transformer models has recently received significant attention from the NLP community.
\citet{sanh2019distilbert} presented DistilBERT, a BERT-like model with 6 layers.
This student was initialized using some of the parameters from a $\textnormal{BERT}_\textsc{Base}$ teacher, and subsequently distilled from it.
\citet{xu2020improving} proposed self-ensemble/distillation methods for BERT models in text classification and NLI tasks; their teachers are obtained by ensembling student models or by averaging of model parameters from previous time steps.
\citet{turc2019well} and \citet{sun2019patient} also explored knowledge distillation for BERT model compression, using smaller BERT models with fewer transformer blocks as student models.
Previous studies on transformer distillation have also leveraged its intermediate representation~\cite{sun2019patient,sun2020mobilebert,jiao2020tinybert,mukherjee2020xtremedistil,liang2020mixkd}.
These approaches typically lead to more accurate performance, but severely limit which pairing of teacher and students can be used (\eg, same transformer family/tokenization, identical hidden dimensions).

\subsection{Ensemble Distillation}
\citet{yang2020model} discussed two-stage multi-teacher knowledge distillation for QA tasks.
Similarly, \citet{jiao2020tinybert} used BERT models as teachers for their proposed model, TinyBERT, in a two-stage learning strategy.
Unlike their two-stage approach, our study focuses on distilling the knowledge of multiple teachers \textit{while} preserving the individual teacher distributions.
Furthermore, we explore several pretrained transformer models for knowledge distillation instead of focusing on a specific architecture.
More recently, \citet{allen2020towards} formally proved that an ensemble of models of the same family can be distilled into a single model while retaining the same performance of the ensemble;
however, their experiments are exclusively focus on ResNet models for image classification tasks.
\citet{kwon2020adaptive} tried to dynamically select, for each training sample, one among a set of teachers.
These studies focus distillation on models that strictly share the same architecture and training strategy, which we show not achieving the same accuracy as our \ourmodel model.

\subsection{Multi-head Transformers}
To the best of our knowledge, no previous work discusses multi-head transformer models for ranking problems;
however, some related works exist for classification tasks.
TwinBERT~\cite{lu2020twinbert} may be the most similar approach to \ourmodel;
it consists of two multi-layer transformer encoders and a crossing layer to combine their outputs.
While TwinBERT has two \emph{bodies} which share \emph{one classification head}, our model aims at the opposite:
\ourmodel consists of \emph{one shared body} and \emph{multiple ranking heads} for efficient inference and multi-teacher knowledge distillation.
Another similar approach is proposed by \citet{tran2020hydra}.
However, this work exclusively focuses on non-transformer models (ResNet-20 V1 from~\citet{he2016deep}) for image classification tasks, and is evaluated only on small datasets such as MNIST and CIFAR.
Besides the different domain, this approach also focuses on distilling from architecturally similar models (distilling 50 ResNet-20 teacher models into a ResNet-20 student with 50 heads), rather than aiming at cross-model family training to increase diversity.

\section{Methodology}

We build up to introducing \ourmodel by first formalizing the \astwo task (Section~\ref{subsec:training}), and then summarizing typical transformer distillation and ensembling techniques (Section~\ref{subsec:distilled}). 
Finally, details of the \ourmodel approach are explained in Section~\ref{subsec:our_model}.

\subsection{Training Transformer Models for Answer Sentence Selection (\astwo)}
\label{subsec:training}

The \astwo task consists of selecting the correct answer from a set of candidate sentences for a given question.
Like many other ranking problems, it can be formulated as a max element selection task: given a query $q \in Q$ and a set of candidates $A=\{a_1, \cdots,a_n\}$, select $a_j$ that is an optimal element for $q$.
We can model the task as a selector function $\pi: Q \times \mathcal{P}(A) \rightarrow A$, defined as $\pi(q,A) = a_j$, where $\mathcal{P}(A)$ is the powerset of $A$, $j= {\tt argmax}_i \left(p(a_i | q)\right)$, and $p(a_i| q)$ is the probability of $a_i$ to be the required element for $q$.
In this work, we evaluate \ourmodel, as well as all our baselines, as an estimator for $p(a_i | q)$ for the \astwo task.
In the remainder of this work, we formally refer to an estimator by using a uppercase calligraphy letter and a set of model parameters $\Theta$, \emph{e.g.}, $\mathcal{M}_\Theta$.

We fine-tune three models to be used as a teacher $\mathcal{T}_\Theta$: \robertaL, \electraL, and \albertXXL.
The first two share the same architecture, consisting of 24 layers and a hidden dimension of 1,024, while \albertXXL is wider (4,096 hidden units) but shallower (12 layers).
All three models are optimized using cross entropy loss in a point-wise setting, \emph{i.e.}, they are trained to maximize the log likelihood of the binary relevance label for each answer separately.

While approaches that optimize the ranking over multiple samples (such as pair-wise or list-wise methods) could also be used \cite{Bian:2017:CMD:3132847.3133089}, they would not change the overall findings of our study; further, point-wise methods have been shown to achieve competitive performance for transformer models \cite{macavaney:sigir2019-cedr}.

When training models for the \gpd and WikiQA datasets, we follow the \tanda technique introduced by~\citet{garg2020tanda}: models are first fine-tuned on ASNQ to transfer to the QA domain, and then adapted to the target task.

Besides the three teacher models, we also train their equivalent \textsc{Base} version, namely \robertaB, \electraB, and \albertB.
These baselines serve as a useful comparison for measuring the effectiveness of distillation techniques.

\subsection{Distilled Models and Ensembles}
\label{subsec:distilled}

Knowledge distillation (KD), as defined by~\citet{hinton2015distilling}, is a training technique which a larger, more powerful \textit{teacher} model $\mathcal{T}_\Theta$ is used to train a smaller, more efficient model, often dubbed as \textit{student} model $\mathcal{S}_\Theta$.
$\mathcal{S}_\Theta$ is typically trained to minimize the difference between its output distribution and the teacher's.
If labeled data is available, it is often used in conjunction with the teacher output as it often leads to improved performance~\citep{ba2014deep}.
In these cases, we train $\mathcal{S}_\Theta$ using a \textit{soft loss} with respect to its teacher and a \textit{hard loss} with respect to the human-annotated labels.

To distill the three \textsc{Large} models introduced in Section~\ref{subsec:training}, we use the loss formulation from~\citet{hinton2015distilling}, as it performs comparably to other, more recent distillation techniques~\citep{tian2019contrastive}.
Given a pair of input sequence $x$ and the target label $y$, it is defined as follows:
\begin{eqnarray}
\hspace{-2em}    &\mathcal{L}_\textnormal{KD}(x, y) =& \alpha \mathcal{L}_\textnormal{H}(\mathcal{S}_\Theta(x), y) + \nonumber\\
    && \hspace{-2em} (1 - \alpha) \tau^2 \mathcal{L}_\textnormal{S}(\mathcal{S}_\Theta(x), \mathcal{T}_\Theta(x))
    \label{eq:kd_loss}
\end{eqnarray}
\noindent where $\alpha$ and $\tau$ indicate a balancing factor and temperature for distillation, respectively.
We independently tune hyperparameters $\alpha \in \{0.0, 0.1, 0.5, 0.9\}$ and $\tau \in \{1, 3, 5\}$ for each dataset on their respective dev sets.
As previously mentioned, we use cross entropy as hard loss $\mathcal{L}_\textnormal{H}$ for all our experiments.
$\mathcal{L}_\textnormal{S}$ is a soft loss function based on the Kullback-Leibler divergence $\textnormal{KL}(p(x), q(x))$, where $p(x)$ and $q(x)$ are softened-probability distributions of teacher $\mathcal{T}_\Theta$ and student $\mathcal{S}_\Theta$ models for a given  input $x$, that is, $p(x) = [p_{1}(x), \cdots, p_{|C|}(x)]$ and $q(x) = [q_{1}(x), \cdots, q_{|C|}(x)]$ defined as follows:
\begin{eqnarray}
    &p_{c}(x) = \frac{\exp \left( {\mathcal{T}_\Theta(x, c) / \tau} \right)}{\sum_{j \in C} \exp\left( {\mathcal{T}_\Theta(x, j) / \tau} \right)} \\
    &q_{c}(x) = \frac{\exp \left( {\mathcal{S}_\Theta(x, c) / \tau} \right)}{\sum_{j \in C} \exp\left( {\mathcal{S}_\Theta(x, j)/\tau} \right)},
    \label{eq:pq}
\end{eqnarray}
\noindent where $C$ indicates a set of class labels.

Using the technique described above, we distill three \textsc{Large} models into their corresponding \textsc{Base} counterparts: \emph{i.e.}, \albertB from \albertXXL, and so on.
Furthermore, we create an ensemble of \textsc{Base} models by linearly combining their outputs;
hyperparameters for ensembles were tuned by Optuna~\cite{akiba2019optuna}.

Finally, we build another ensemble model of three \electraB distilled from the three \textsc{Large} models mentioned above.
As we will show in Section~\ref{sec:experiments}, \electraB outperforms all other \textsc{Base} models;
therefore, we are interested in measuring whether it could be used for inter transformer family model distillation.
Once again, Optuna was used to tune the ensemble model.

We note that the ensemble of the three \textsc{Large} models is not used as a teacher.
In our preliminary experiment, we found that the ensemble is not a good teacher, as the model was \emph{too confident} in its prediction, a trend that is studied by \citet{panagiotatos2019curriculum}.
Most softmaxed category-probabilities by the ensemble model are close to either 0 or 1 and behave like hard-target rather than soft-target, which did not improve over the KD baselines (rows 7--9) in Table~\ref{table:performance}.

\subsection{\ourmodel: Multiple-Heads Student}
\label{subsec:our_model}

\begin{figure}[tb]
    \centering
    \includegraphics[width=0.95\linewidth]{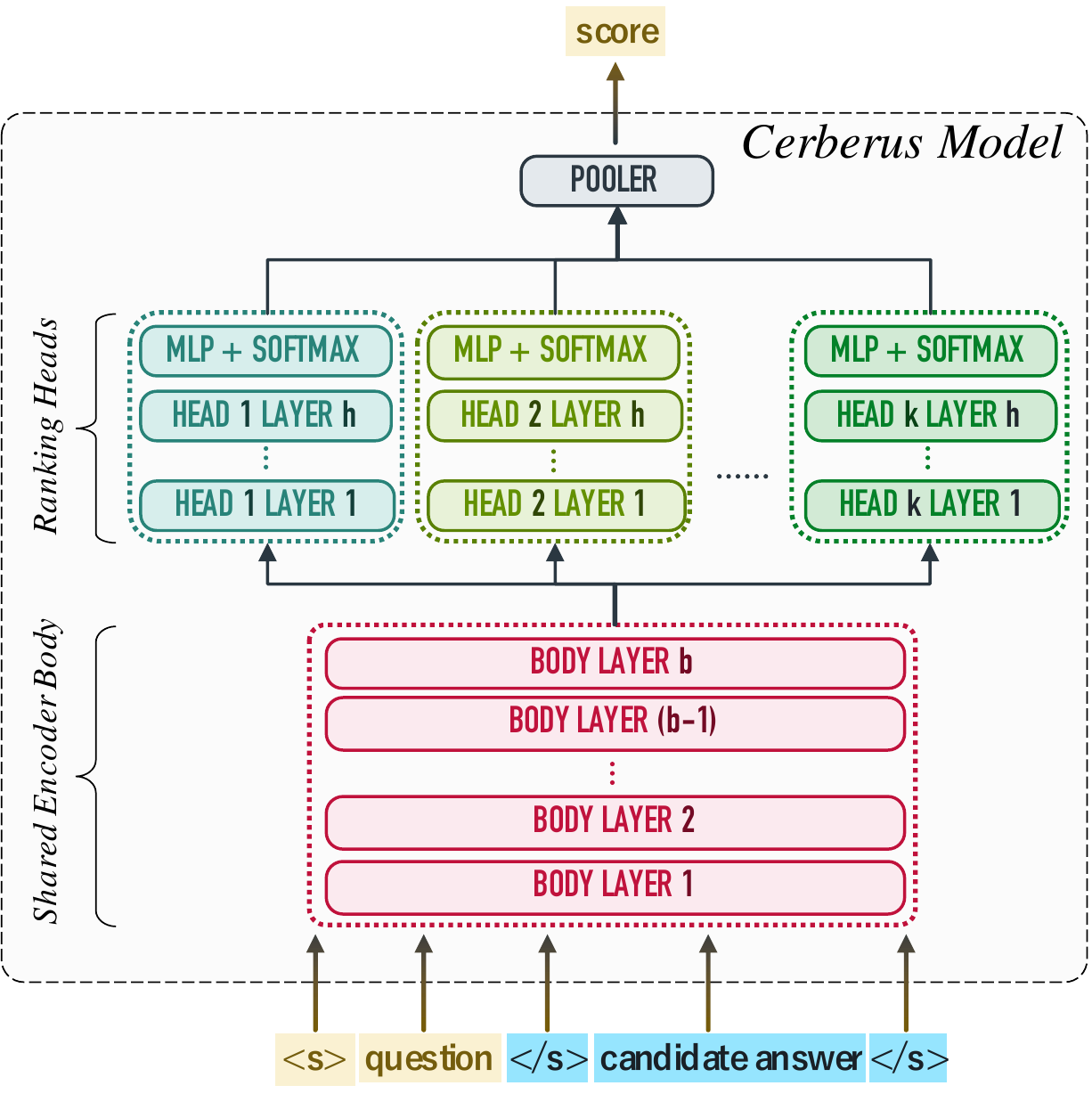}
    \caption{
        Detailed overview of \ourmodel model that consists of a shared encoder body of $b$ transformer layers, followed by $k$ ranking heads of $h$ layers each; we use notation $\mathit{B}_b\ k\textit{H}_{h}$ to identify a \ourmodel configuration.
        {\bf All heads are jointly trained, but each head learns from a unique teacher model}; at inference time, predictions from heads are combined by a pooler layer.
    }
    \label{fig:our_model_design}
\vspace{-1em}
\end{figure}

As mentioned in the previous section, students trained using different teachers can be trivially ensembled using a linear combination of their outputs.
However, this results in a drastic increase in model size, as well as a synchronization latency overhead, which are both undesirable properties in many applications.
In this section, we introduce \ourmodel, a transformer architecture designed to emulate the properties of an ensemble of distilled models while being more efficient.
As illustrated in Fig.~\ref{fig:our_model_design}, our \ourmodel model consists of two components: (\textit{i}) an input encoder comprised of stacked transformer layers, and (\textit{ii}) a set of $k$ ranking heads, each designed to be trained with respect to a specific teacher.
Each ranking head is comprised of one or more transformer layers;
it receives as input the output of the shared encoder, and produces classification output.
To obtain its final prediction, the \ourmodel averages the outputs of its ranking heads.
A schematic representation of \ourmodel is shown in Figure~\ref{fig:our_model_design}.

Formally, let $M_\Theta$ be a pretrained transformer\footnote{In our experiments on ASNQ, we use a pretrained \electraB model as starting point; for \gpd and WikiQA, we use a \electraB model fine-tuned on ASNQ.} of $n$ layers.
To obtain a \ourmodel model, we first split the model into two groups: the first $b$ blocks are used for the shared encoder body $B_b$, while the next $h=(n-b)$ blocks are replicated and assigned as initial states for each head $H_h^i$, $i=\{1,\ldots,k\}$.
To compute the output for the $i^\textnormal{th}$ head, we first encode an input $x$ using $B_b$, and then use it as input to $H_h^i$.
To train \ourmodel, we use a linear combination of $k$ loss functions, each of which uses output from a different ranking head:
\begin{equation}
    \mathcal{L}_\text{\ourmodel}(x, y) = \textstyle\sum_{i=1}^{k} \lambda_{i} \cdot \mathcal{L}_{i}(x, y),
    \label{eq:our_model_loss}
\end{equation}
\noindent where $\lambda_{i}$ and $\mathcal{L}_{i}$ are the weight and loss function for the $i$-th head in the \ourmodel model.
Specifically, we apply the loss function of Equation~\ref{eq:kd_loss} to each head, \emph{i.e.}, $\mathcal{L}_{i} = \mathcal{L}_\textnormal{KD}$ for the $i^\textnormal{th}$ head-teacher pair.
We note that, while the encoder body and all ranking heads are trained jointly, each head is optimized only by its own loss.
Conversely, when backpropagating $\mathcal{L}_\textnormal{\ourmodel}$, the parameters of the encoder body are affected by the output of all $k$ ranking heads.
This ensures that each head learns faithfully from their teacher while the parameters of the encoder body remain suitable for the entire model.

For inference, a single score for \ourmodel is obtained by averaging the outputs of all ranking heads:
\vspace{-0.25em}
\begin{equation}
    \textnormal{score}(x) = \frac{1}{k}\textstyle\sum_{i=1}^{k} H^i_h(B_b(x)).
    \label{eq:our_model_output}
\vspace{-0.25em}
\end{equation}

In our experiments, we use $k=3$ heads, each trained with one of the \textsc{Large} models described in Section~\ref{subsec:training}.
We discuss a variety of combination for values of $b$ and $h$;
the performance for each configuration is analyzed in Section~\ref{subsec:how_deep}.
For training, we set $\lambda_{i}=1$ for all $i=\{1,\ldots,k\}$ and reuse the search space of the hyperparameters $\alpha$ and $\tau$ for knowledge distillation (see Section~\ref{subsec:distilled}).

\section{Experimental Setup}
\label{sec:experiments}

\subsection{Datasets}

\begin{table}[t]
    \centering
    \small
	\bgroup
    \setlength{\tabcolsep}{0.4em}
	\renewcommand{\arraystretch}{1.15}
    \begin{tabular}[t]{llrrr}
        \toprule
        &
        & {\textbf{ASNQ}}
        & { \textbf{\gpd}}
        & {\textbf{WikiQA}}	\\
        \midrule
        \multirow{3}{*}{\STAB{\rotatebox[origin=c]{90}{\textsc{train}}}}
        & Questions			    & 57,242 		& 3,074			& 873				\\
        & QA pairs	    & 23,662,238  		& 189,050			& 8,672				\\
        & Correct answers  	& 69,002 	 		& 32,284		& 1,040				\\
        \midrule
        \multirow{3}{*}{\STAB{\rotatebox[origin=c]{90}{\textsc{dev}}}}
        & Questions			    & 1,336			& 808			& 126				\\
        & QA pairs  	& 539,210			& 20,135			& 1,130				\\
        & Correct answers  	& 4,166			& 5,945			& 140				\\
        \midrule
        \multirow{3}{*}{\STAB{\rotatebox[origin=c]{90}{\textsc{test}}}}
        & Questions 			& 1,336			& 3,000			& 243				\\
        & QA pairs  	& 535,116			& 74,670			& 2,351				\\
        & Correct answers  	& 4,250			& 21,328			& 293				\\
        \bottomrule
    \end{tabular}
    \egroup
    \caption{Dataset statistics. ASNQ and \gpd contain significantly more candidates than WikiQA.}
    \label{tab:dataset_stats}
\end{table}

While many studies on Transformer-based models~\cite{devlin2019bert,liu2019roberta,clark2019electra,lan2019albert} are assessed for GLUE tasks (10 classification and 1 regression tasks), our interests are in ranking problems for question answering such as \astwo.
To fairly assess the \astwo performance of our proposed method against conventional distillation techniques, we report experimental results on a set of three diverse English \astwo datasets: 
WikiQA~\cite{yang2015wikiqa}, a small academic dataset that has been widely used;
ASNQ~\cite{garg2020tanda}, a much larger corpus (3 orders of magnitude larger than WikiQA) that allow us to assess models' performance in data-unbalanced settings;
finally, we measure performance on \gpd, an internal dataset we constructed for \astwo. 
Compared to the other two corpora, \gpd contains noisier data and is much closer to a real-world \astwo setting. 
Table~\ref{tab:dataset_stats} reports the statistics of the datasets, and more details are described in Appendix.

\begin{table*}[th]
	\centering
	\bgroup
	\small
    \setlength{\tabcolsep}{0.4em}
	\renewcommand{\arraystretch}{1.5}
	\begin{tabular}{r@{\hspace{2\tabcolsep}}c@{\hspace{1\tabcolsep}}c@{\hspace{1\tabcolsep}}r@{\hspace{3\tabcolsep}}rrr@{\hspace{3\tabcolsep}}rrr@{\hspace{3\tabcolsep}}rrr}
	\toprule
	\multirow{2}{*}{} & \multirow{2}{*}{\normalsize\textbf{Teacher}} & \multirow{2}{*}{\normalsize\textbf{Student}} & \multirow{2}{*}{\normalsize
    \renewcommand{\arraystretch}{1.01}\begin{tabular}[c]{@{}c@{}} \textbf{Params} \\ \textbf{count} \end{tabular}
	} & \multicolumn{3}{c}{\normalsize\textbf{\gpd}} & \multicolumn{3}{c}{\normalsize\textbf{ASNQ}} & \multicolumn{3}{c}{\normalsize\textbf{WikiQA}} \\
	 &  &  &  & \multicolumn{1}{l}{\textbf{P@1}} & \multicolumn{1}{l}{\textbf{MAP}} & \multicolumn{1}{l}{\textbf{MRR}} & \multicolumn{1}{l}{\textbf{P@1}} & \multicolumn{1}{l}{\textbf{MAP}} & \multicolumn{1}{l}{\textbf{MRR}} & \multicolumn{1}{l}{\textbf{P@1}} & \multicolumn{1}{l}{\textbf{MAP}} & \multicolumn{1}{l}{\textbf{MRR}} \\
	\midrule
	{\footnotesize \color{gray} \it  1} & N/A & \albertXXL & 222M & {63.9} & {60.5} & {69.9} & 60.8 & 72.7 & 72.6 & 87.2 & 91.4 & 92.6 \\
	{\footnotesize \color{gray} \it  2} & N/A & {\renewcommand{\arraystretch}{1.01}\begin{tabular}[c]{@{}c@{}} \robertaL \\ {\fontsize{7pt}{7pt}\selectfont \citep{garg2020tanda}} \end{tabular}} & 335M & 64.2 & 60.6 & \thirdresult{70.3} & 62.6 & 73.6 & 73.7 & \firstresult{89.3} & \firstresult{92.6} & \firstresult{93.6} \\
	{\footnotesize \color{gray} \it  3} & N/A & \electraL & 335M & \secondresult{65.0} & \firstresult{61.3} & \secondresult{70.7} & \firstresult{64.7} & \secondresult{74.4} & \secondresult{74.7} & 86.7 & 91.0 & 92.2 \\
	\midrule
	{\footnotesize \color{gray} \it  4} & N/A & \albertB & 11M & 58.8 & 55.6 & 66.1 & 49.3 & 63.2 & 62.5 & 83.5 & 88.9 & 90.1 \\
	{\footnotesize \color{gray} \it  5} & N/A & {\renewcommand{\arraystretch}{1.01}\begin{tabular}[c]{@{}c@{}} \robertaB \\ { \fontsize{7pt}{7pt}\selectfont\citep{garg2020tanda}} \end{tabular}} & 109M & 59.6 & 56.6 & 67.0 & 54.9 & 67.2 & 67.0 & 82.7 & 88.7 & 89.8 \\
	{\footnotesize \color{gray} \it  6} & N/A & \electraB & 109M & 62.2 & 58.8 & 68.7 & 61.8 & 71.9 & 72.3 & 86.3 & 90.7 & 91.9 \\
	\midrule
	{\footnotesize \color{gray} \it  7} & \albertXXL & \albertB & 11M & 61.5 & 57.2 & 68.0 & 56.5 & 68.5 & 68.6 & 84.0 & 89.0 & 90.3 \\
	{\footnotesize \color{gray} \it  8} & \robertaL & \robertaB & 109M & 63.4 & 59.4 & 69.7 & 62.4 & 72.2 & 72.6 & 83.5 & 89.2 & 90.6 \\
	{\footnotesize \color{gray} \it  9} & \electraL & \electraB & 109M & 63.2 & \secondresult{61.1} & 69.6 & \thirdresult{63.7} & \thirdresult{73.9} & \thirdresult{74.1} & 88.1 & 91.6 & \thirdresult{92.9} \\
	\midrule
	{\footnotesize \color{gray} \it  10} & \multicolumn{2}{c}{Ensemble of 3 \textsc{base} (rows 4--6)} & 247M & 63.7 & 59.5 & 69.6 & 62.2 & 72.5 & 72.9 & 88.1 & 91.4 & 92.7 \\
	{\footnotesize \color{gray} \it  11} & \multicolumn{2}{c}{Ensemble of 3 distilled (rows 7--9)}& 247M & 64.2 & 60.0 & 70.1 & 62.7 & 72.8 & 73.1 & 88.1 & \thirdresult{91.7} & \thirdresult{92.9} \\
	{\footnotesize \color{gray} \it  12} & \multicolumn{2}{c}{\renewcommand{\arraystretch}{1.01}\begin{tabular}[c]{@{}c@{}}$\textnormal{Ensemble}$ of 3 \electraB\\distilled from $\ast_\textsc{Large}$ (rows 1--3)\end{tabular}} & 327M & \firstresult{65.1} & \thirdresult{60.8} & \firstresult{70.8} & {63.6} & {73.5} & {74.0} & \thirdresult{88.6} & {91.5} & {92.8} \\
	\midrule
	{\footnotesize \color{gray} \it  13} & \multicolumn{2}{c}{Hydra \citep{tran2020hydra}} & 124M & 63.4 & 59.9 & 69.7 & 62.7 & 73.0 & 73.3 & 88.1 & 91.5 & 92.8 \\
	\midrule
    {\footnotesize \color{gray} \it  14} & \multicolumn{1}{c}{\renewcommand{\arraystretch}{0.95}\begin{tabular}[c]{@{}c@{}}$\ast_\textsc{Large}$ \\(rows 1--3)\end{tabular}} & {\renewcommand{\arraystretch}{1.01}\begin{tabular}[c]{@{}c@{}}$\text{\ourmodel}_{B_{11} 3H_1}$\\{\fontsize{8pt}{8pt}\selectfont
    {\textbf{(our approach)}}}~\end{tabular}} & 124M & \thirdresult{64.3} & \thirdresult{60.8} & \thirdresult{70.3} & \secondresult{64.3} & \firstresult{75.1} & \firstresult{75.2} & \firstresult{89.3} & \secondresult{92.4} & \secondresult{93.5} \\

	\bottomrule
	\end{tabular}
	\egroup
	\vspace{-0.5em}
	\caption{Performance on \gpd, ASNQ, and WikiQA.
	For each metric, we highlight the \firstresult{best}, \secondresult{$\textnormal{2}^\textnormal{nd}$ best}, and \thirdresult{$\textnormal{3}^\textnormal{rd}$ best} scores.
	We compare our \ourmodel model (row 14) with state-of-the-art \astwo models (\citet{garg2020tanda}, rows 2 and 5), ensembles from distilled models (rows 10--12), and the technique proposed by \citet{tran2020hydra} (row 13). 
    \ourmodel achieves equivalent performance (Spearman~$\rho$, $p < 0.01$) of state-of-the-art AS2 models while $2.7\times$ smaller; it outperforms all models with a comparable number of parameters (Wilcoxon signed-rank test, $p < 0.01$).
	}
	\label{table:performance}
	\vspace{-.5em}
\end{table*}

\subsection{Evaluation Metrics}
\label{subsec:metrics}

We assess \astwo performance on ASNQ, WikiQA and \gpd using three metrics: mean average precision (MAP), mean reciprocal rank (MRR), and precision at top-1 candidate (P@1).
The first two metrics are commonly used to measure overall performance of ranking systems, while P@1 is a stricter metric that captures effectiveness of high-precision applications such as \astwo.

Our models are implemented with PyTorch 1.6~\cite{pytorch} using Hugging Face Transformers 3.0.2~\cite{wolf2020huggingface};
all models are trained on a machine with 4 NVIDIA Tesla V100 GPUs, each with 16GB of memory. 
Latency benchmarks are executed on a single GPU to eliminate variability due to inter-accelerator communication.

\section{Results}
\label{sec:results}

Here we present our main experimental findings.
In Section~\ref{subsec:as2_results}, we compare \ourmodel to state-of-the-art models and other distillation techniques using three datasets (\gpd, ASNQ, WikiQA).
In Sections~\ref{subsec:why_multiple_heads} -- \ref{subsec:how_deep}, we motivate our design and hyperparameter choices for \ourmodel by empirically validating them.
Finally, in Section~\ref{subsec:inference_time}, we discuss inference latency of \ourmodel comparing to other transformer models. 

\subsection{Answer Sentence Selection Performance}
\label{subsec:as2_results}

The performance of \ourmodel on \gpd, ASNQ, and WikiQA datasets are reported in Table~\ref{table:performance}. 
Specifically, we compared our approach (row 14) to four groups of baselines: 
larger transformer-based models (rows 1--3), including the state-of-the-art \astwo models by \citet{garg2020tanda} (rows 2 and 5); 
equivalently sized models, either directly fine-tuned on target datasets (rows 4--6), or distilled using their corresponding \textsc{Large} model as teacher (rows 7--9);
ensembles of \textsc{Base} models (rows 10--12).
We also adapted the ensembling technique of Hydra~\cite{tran2020hydra}, which is originally designed for image recognition, to work in our \astwo setting\footnote{Instead of 50 ResNet-20 V1 teachers paired with a 50-head ResNet-20 V1 student, we train 3 \electraB teachers with different seeds and distill them into a \ourmodel model (referred to as Hydra in Table~\ref{table:performance}) initialized from \electraB.\label{fn:hydra}} and used it as a baseline (row 13). 
All the comparisons are done with respect to a $B_{11} 3H_1$ \ourmodel model initialized from an \electraB model: 
performance of other model configurations are discussed in Section~\ref{subsec:how_deep}.
Due to the volume of experiments, we train a model with a random seed for each model given a set of hyperparameters and report the \astwo performance with the best hyperparameter set according to each dev set.

\subsubsection{Vs. \tanda (\textsc{Base}) \& Single-Model Distillation}
We find \textsc{Base} models trained by \tanda (rows 4--6), the state-of-the-art training method for \astwo tasks, are further improved (rows 7--9) by introducing knowledge distillation to its 2nd fine-tuning stage.
Our \ourmodel achieves a significantly improvement over all single \textsc{Base} models for all the considered datasets (Wilcoxon signed-rank test, $p < 0.01$). 
We empirically show in Section~\ref{subsec:why_multiple_heads} that this significant improvement was achieved by both the architecture of our \ourmodel and using heterogeneous teacher models rather than a small amount of extra parameters.

\subsubsection{Vs. \tanda (\textsc{Large})}
We observe that our \ourmodel equals (Spearman's rank correlation, $p < 0.01$) \textsc{LARGE} models trained by \tanda (rows 1--3), including the state-of-the-art \astwo model~\cite{garg2020tanda} (row 2), while the \ourmodel has 2.7 times fewer parameters.
Furthermore, our \ourmodel consistently outperforms \albertXXL, which has 1.8 times more parameters than the \ourmodel.

\subsubsection{Vs. Ensembles \& Hydra}
For all the datasets we considered, our \ourmodel achieves similar or better performance of much larger ensemble models, including an ensemble of \albertB, \robertaB, and \electraB trained with and without distillation (rows 10 and 11), as well as the ensemble of three \electraB models each trained using \albertXXL, \robertaL, and \electraL as teachers (row 12).
We also note that \ourmodel outperforms our adaptation of Hydra~\cite{tran2020hydra} (row 13), which emphasizes the importance of using heterogeneous teacher models for \astwo.

\begin{table}[t]
\centering
\bgroup
\small
\setlength{\tabcolsep}{0.4em}
\renewcommand{\arraystretch}{1.3}
\begin{tabular}{lcrrr} 
\toprule
\multicolumn{1}{l}{
\renewcommand{\arraystretch}{.95}\begin{tabular}[c]{@{}l@{}} {\bf Distillation} \\ {\bf Strategy} \end{tabular}
} & \multicolumn{1}{c}{\bf Model} & \multicolumn{1}{c}{\textbf{P@1}} & \multicolumn{1}{c}{\textbf{MAP}} & \multicolumn{1}{c}{\textbf{MRR}} \\ 
\midrule
\multicolumn{1}{l}{
\renewcommand{\arraystretch}{.95}\begin{tabular}[c]{@{}l@{}} \tanda \\ (\text{no teacher}) \end{tabular}
} &\electraB & 62.2 & 58.8 & 68.7  \\
Single teacher &\electraB & 63.2 & 61.1 & 69.6  \\
$\textnormal{KD}_\textnormal{Sum}$ & \electraB & 63.5 & 60.1 & 69.8  \\
$\textnormal{KD}_\textnormal{RR}$ & \electraB & 63.1 & 60.2 & 69.6 \\
\midrule
\multicolumn{1}{l}{
\renewcommand{\arraystretch}{.95}\begin{tabular}[c]{@{}l@{}} \tanda \\ (\text{no teacher}) \end{tabular}
} & $\text{\ourmodel}_{B_{11} 3H_1}$ & 62.1 & 59.5 & 68.8  \\
Single teacher & $\text{\ourmodel}_{B_{11} 3H_1}$ & 63.2 & 60.5 & 69.5 \\
Hydra\footref{fn:hydra} & $\text{\ourmodel}_{B_{11} 3H_1}$ & 63.4 & 59.9 & 69.7 \\
\midrule
\multicolumn{1}{l}{
\renewcommand{\arraystretch}{.95}\begin{tabular}[c]{@{}l@{}} One teacher \\ per head \end{tabular}
} & $\text{\ourmodel}_{B_{11} 3H_1}$ & \textbf{64.3} & \textbf{60.8} & \textbf{70.3} \\
\bottomrule
\end{tabular}
\egroup
\caption{Comparison of single and multiple teachers distillation for \electraB and \ourmodel $B_{11} 3H_1$ models on the \gpd test set. 
Overall, we found that combining the \ourmodel architecture with multiple teachers is essential to achieve the best performance.}
\label{table:why_multiple_heads}
\end{table}

\subsection{Are Multiple Ranking Heads and Heterogeneous Teachers Necessary?}
\label{subsec:why_multiple_heads}
Using the heterogeneous teacher models shown in Table~\ref{table:performance}, we discuss how \astwo performance varies when using different combinations of teachers for knowledge distillation.
The first method, $\textnormal{KD}_\textnormal{Sum}$, simply combines loss values from multiple teachers to train a single transformer model, similarly to the task-specific distillation stage with multiple teachers in~\citet{yang2020model}.
In the second method, $\textnormal{KD}_\textnormal{RR}$, we switch teacher models for each training batch in a round-robin style; 
\emph{i.e.}, the student transformer model will be trained with the first teacher model in the first batch, with the second teacher model in the second batch, and so forth.

Table~\ref{table:why_multiple_heads} compares performance of the multiple-teacher knowledge distillation strategies described above to that of our proposed method; we also evaluate the effect of using one teacher per head, rather than a single teacher (\electraL), on \ourmodel. 
For \electraB, we found that $\textnormal{KD}_\textnormal{Sum}$ method slightly outperforms $\textnormal{KD}_\textnormal{RR}$;
this result highlights the importance of leveraging multiple teachers for knowledge distillation in the same mini-batch.
For \ourmodel, we found that using multiple heterogeneous teachers (specifically, one per ranking head) is crucial in achieving the best performance;
without it, \ourmodel $B_{11} 3H_1$ achieves the same performance of \electraB despite having more parameters.
Besides these two trends, the results of rows 13 and 14 in Table~\ref{table:performance} emphasize the importance of heterogeneity in the set of teacher models.

As a result, \ourmodel $B_{11} 3H_1$ performs the best and achieves the comparable performance with some of the teacher (\textsc{Large}) models, while saving between 45\% and 63\% of model parameters.
From the aforementioned three trends, we can confirm that the improved \astwo performance was achieved thanks to the \textbf{multiple ranking heads} in \ourmodel, the use of \textbf{multiple teachers}, and \textbf{heterogeneity} in teacher model families; on the other hand, the slightly increased parameters compared to \electraB did not contribute to performance uplift.

\subsection{Do Heads Resemble Their Teachers?}
\label{subsec:like_teacher}

\begin{figure}[t]
    \centering
    \includegraphics[width=0.875\linewidth]{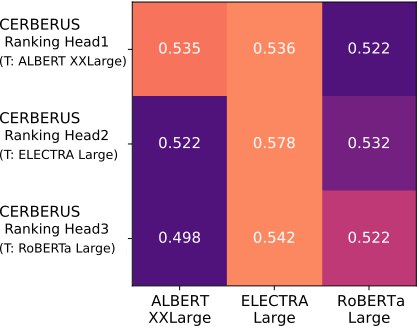}
    \caption{
        Agreement between heads and their teacher model in \ourmodel. 
        It is obtained by diving the number of correct candidates each head and teacher agree on by the total number of correct answer for each head. 
    }
    \label{fig:agreements}
\end{figure}

To better understand the relationship between \ourmodel's ranking heads and the teachers used to train them, we analyze the top candidates chosen by each of teacher and student models.
Figure~\ref{fig:agreements} shows how often each \ourmodel head agrees with its respective teacher model. 
To calculate agreement, we normalize number of correct candidates heads and teachers agree on by the total number of correct answer for each head.

Intuitively, we might expect that ranking heads would agree the most with their respective teachers;
however, in practice, we notice that the highest agreement for all heads is measured with  \electraL.
However, one should consider that the agreement measurement is confounded by the fact that all heads are more likely to agree with the head that is correct the most (\electraL).
Furthermore, in all our experiments, \ourmodel is initialized from a pretrained \electraB, which also increase the likelihood of agreement with \electraL. 
Nevertheless, we do note that both the head distilled from \albertXXL and from \robertaB achieve high agreement with their teachers, suggesting that \ourmodel ranking heads do indeed resemble their teachers.

In our experiments, we also observed that \ourmodel is able to mimic the behavior of an ensemble comprised of the three large models; for example, on the WikiQA dataset, \ourmodel always predicts the correct label when all three models are correct (197/243 queries), it follows majority voting in 17 cases, and in one case it overrides the majority voting when one of the teachers is very confident.
In the remaining cases, either only a minority or no teachers are correct, or the confidence of the majority is low.

\subsection{How Many Blocks Should Heads Have?}
\label{subsec:how_deep}

In Table~\ref{table:performance}, we examined the performance of a \ourmodel model with configuration $B_{11} 3H_1$; 
that is, a body composed of 11 blocks, and 3 ranking heads with one transformer block each.
In order to understand  how specific hyperparameters setting for \ourmodel influences model performance, we examine different \ourmodel configurations in this section. 
Due to space constraints, we only report results on \gpd; we observed similar trends on ASNQ and \gpd.
In order to keep a latency comparable to that of other \textsc{Base} models, we keep the total depth of \ourmodel constant, and vary the number of blocks in the ranking heads and shared encoder body. 

Table~\ref{table:our_model_size} shows the results for alternative \ourmodel configurations. 
Overall, we noticed that the performance is not significantly affected by the specific configuration of \ourmodel, which yields consistent results regardless of the number of transformer layers used (1 to 6, $B_{11} 3H_1$ to $B_{6} 3H_6$). 
All \ourmodel models are trained with a combination of hard and soft losses, which makes it more likely to have different configurations converge on a set of stable but similar configurations.
Despite the similar performance, we note that $B_{6} 3H_6$ is comprised of significantly more parameters than our leanest configuration, $B_{11} 3H_1$ (199M vs 124M). 
Given the lack of improvement from the additional parametrization, all experiments in this work were conducted by using 11 shared body blocks and 3 heads, each of which consists 1 block ($B_{11} 3H_1$).

\begin{table}[t]
\centering
\bgroup
\small
\setlength{\tabcolsep}{0.8em}
\renewcommand{\arraystretch}{1.3}
\begin{tabular}{ccrrr} 
\toprule
\multicolumn{1}{c}{
\renewcommand{\arraystretch}{1.01}\begin{tabular}[c]{@{}c@{}} {\bf \ourmodel} \\ {\bf config} \end{tabular}
} & \multicolumn{1}{c}{
\renewcommand{\arraystretch}{1.01}\begin{tabular}[c]{@{}c@{}} {\bf Params} \\ {\bf count} \end{tabular}
} & \multicolumn{1}{c}{\textbf{P@1}} & \multicolumn{1}{c}{\textbf{MAP}} & \multicolumn{1}{c}{\textbf{MRR}} \\ 
\midrule
$B_{11}~~3H_1$ & 124M & 64.3 & 60.8 & 70.3 \\
$B_{10}~~3H_2$ & 139M & \textbf{64.9} & \textbf{61.2} & \textbf{70.8} \\
$B_{8~~}~~3H_4$ & 169M & 64.7 & 60.7 & 70.4 \\
$B_{6~~}~~3H_6$ & 199M & 64.3 & 60.6 & 70.3 \\
\bottomrule
\end{tabular}
\egroup
\caption{
Performance of different \ourmodel configurations on the \gpd test set.
Overall, we found that \ourmodel is stable with respect to the configuration. 
}
\label{table:our_model_size}
\end{table}

\subsection{Benchmarking Inference Latency}
\label{subsec:inference_time}

Besides \astwo performance, we examine the inference latency for \ourmodel and models evaluated in Section~\ref{subsec:as2_results}, using an NVIDIA Tesla V100 GPU.
The results are summarized in Table~\ref{table:inf_speed}. 
For a fair comparison between the models, we used the same batch size (128) for all benchmarks, and ignored any tokenization and CPU/GPU communication overhead while recording wall clock time. 
Overall, we confirm that \ourmodel achieves a comparable latency of other \textsc{Base} models. 
All four are within the one standard deviation of each other.

\begin{table}[t]
\centering
\bgroup
\small
\setlength{\tabcolsep}{0.7em}
\renewcommand{\arraystretch}{1.3}
\begin{tabular}{crrr} 
\toprule
\multirow{2}{*}{\textbf{Model} } & \multicolumn{1}{c}{\multirow{2}{*}{
    \renewcommand{\arraystretch}{1.01}\begin{tabular}[c]{@{}c@{}} {\bf Params} \\ {\bf count} \end{tabular}
}} & \multicolumn{2}{c}{\textbf{Latency ($\mathbf{{\mu}s}$)} } \\
 & \multicolumn{1}{c}{} & \multicolumn{1}{c}{\textbf{\textit{avg}}} & \multicolumn{1}{c}{\textbf{\textit{std}}} \\
\midrule
\albertB & 11M & 2.3 & 0.017 \\
\robertaB & 109M & \textbf{1.9} & 0.017\\
\electraB & 109M & 2.0 & 0.018 \\
\albertXXL & 222M & 47.0 & 0.370 \\
\robertaL & 335M & 6.5 & 0.066 \\
\electraL\vspace{.2em} & 335M & 6.7 & 0.089 \\
\multicolumn{1}{c}{
    \renewcommand{\arraystretch}{.95}
    \begin{tabular}[c]{@{}c@{}} 
    Ensemble of \textsc{Base} \\ 
    {\it \color{gray} (Rows 10--11 in Table~\ref{table:performance})}
    \vspace{.2em}
    \end{tabular}
} & 247M & 6.3 & 0.060 \\
\multicolumn{1}{c}{
    \renewcommand{\arraystretch}{.95}
    \begin{tabular}[c]{@{}c@{}} 
    Ensemble of 3 \\ 
    \electraB
    \vspace{.2em}
    \end{tabular}
}& 327M & 6.1 & 0.064 \\
$\text{\ourmodel}_{B_{11} 3H_1}$ & 124M & 2.6 & 0.020 \\
\bottomrule
\end{tabular}
\egroup
\caption{Inference latency.
For a fair comparison, batch size is set to 128 for all models.
\ourmodel achieves latency similar to those of other \textsc{Base} models.}
\label{table:inf_speed}
\end{table}

All the \textsc{Large} models including the state-of-the-art AS2 model (\robertaL by \citet{garg2020tanda}) produce significantly higher latency, (on average, $3.4\times$ slower than \ourmodel);
specifically, \albertXXL, which is comprised of 12 very wide transformer blocks, shows the worst latency among single models.
Further, the latency of the two ensemble models are comparable to some of the \textsc{Large} models, thus supporting our argument that they are not suitable for high performance applications. 
On the other hand, our \ourmodel model saves up to 59\% latency and 62\% model size compared to the ensemble model, while achieving comparable \astwo performance.

\section{Conclusions and Future Work}

In this work, we introduce a technique for obtaining a single efficient \astwo model from an ensemble of heterogeneous transformer models. 
This efficient approach, which we call \ourmodel, consists of a sequence of transformer blocks, followed by multiple ranking heads;
each head is trained with a unique teacher, ensuring proper distillation of the ensemble. 
Results show that the proposed model outperforms traditional, single teacher techniques, rivaling state-of-the-art \astwo models while saving 64\% and 60\% in model size and latency, respectively.
\ourmodel enables \textsc{Large}-like \astwo accuracy while maintaining \textsc{Base}-like efficiency. 

Further analysis demonstrates that reported improvements in \astwo performance are due to to three key factors: (\textit{i}) multiple ranking heads, (\textit{ii}) multiple teachers, and (\textit{iii}) heterogeneity in teacher models.

Future work would focus on two key aspects: how \ourmodel performs on non-ranking tasks, and whether it could achieve similar improvements in ranking tasks outside QA. 
For the former, we remark that, while the core idea of \ourmodel can be extended to tasks such as those in the GLUE benchmark~\citep{glue}, further investigation is necessary in establishing the best set of trade-offs for different objectives and metrics. 
A similar concern exists in the case of extending \ourmodel to ranking tasks, such as ad-hoc retrieval. 

\section*{Limitations}
In this study, we discussed the experimental results and empirically showed the effectiveness of our proposed approach for English datasets only.
While this is a major limitation of the study, our approach is not specific to English, thus it could be extended in the future using models in other languages, although improvements might not translate to less resource-rich languages. 

As described in Section~\ref{subsec:metrics}, our experiments are compute-intensive and have been conducted on 4 NVIDIA V100 GPUs. 
Thus, researchers with less compute might not be able to replicate \ourmodel.

Next, all models we present in this work are trained to optimize answer relevance to a given question. 
Therefore, they might be unfair towards protected categories (race, gender, sex, nationality, etc.) or present answers from a biased point of view. 
Our work does not address this challenge.

Finally, we evaluated our approach only in the context of answer sentence ranking; 
thus, the reader might be left wondering whether such an approach would work for other tasks.
We note that, although a study on the general applicability of our approach is very interesting and needed, it would require more space than a conference submission has in order to be accurately described and evaluated.
Therefore, we leave further investigation of \ourmodel on other domains and tasks as future work. 

\bibliographystyle{acl_natbib}
\bibliography{references}

\newpage
\appendix

\section{WikiQA}
This dataset was introduced by~\citet{yang2015wikiqa} and consists of 1,231 questions and 12,139 candidate answers. 
It was created using queries submitted by Bing search engine users between May 1st, 2010 and July 31st, 2011. 
The dataset includes queries that start with a \textit{wh-} word and end with a question mark. 
Candidates consist of sentences extracted from the first paragraph from Wikipedia page retrieved for each question; they were annotated using Mechanical Turk workers. 
Some of the questions in WikiQA have no correct answer candidate; following \citep{wang2017compare,garg2020tanda}, we remove them from the training set, but leave them in the development and test sets.

\section{ASNQ}
\citet{garg2020tanda} introduced Answer Sentence Natural Questions, a large-scale answer sentence selection dataset. 
It was derived from the Google Natural Questions (NQ)~\citep{kwiatkowski2019nq}, and contains over 57k questions and 23M answer candidates. 
Its large-scale (at least two orders of magnitude larger than any other \astwo dataset) and class imbalance (approximately one correct answer every 400 candidates) properties make it particularly suitable to evaluate how well our models generalize. 
Samples in Google NQ consist of tuples $\langle\texttt{question},$ $\texttt{answer}_\texttt{long},$ $\texttt{answer}_\texttt{short},$ $\texttt{label}\rangle$, where $\texttt{answer}_\texttt{long}$ contains multiple sentences, $\texttt{answer}_\texttt{short}$ is fragment of a sentence, and \texttt{label} indicates whether $\texttt{answer}_\texttt{long}$ is correct.
Google NQ has long and short answers for each question.
To construct ASNQ, \citet{garg2020tanda} labeled any sentence from $\texttt{answer}_\texttt{long}$ that contains $\texttt{answer}_\texttt{short}$ as positive; all other sentences are labeled as negative.
The original release of ANSQ only contains train and development splits; 
We use the dev and test splits introduced by~\citet{soldaini2020cascade}.

\section{\gpd} 
This is an in-house dataset, called \gpdfull, we built as part of our efforts of understanding and benchmarking web-based question answering systems.
To obtain questions, we first collected a non-representative sample of queries from traffic log of our commercial virtual assistant system. 
We then used a retrieval system containing hundreds of million of web pages to obtain up to 100 web pages for each question. 
From the set of retrieved documents, we extracted all candidate sentences and ranked them using \astwo models trained by TANDA~\citet{garg2020tanda};
at least top-25 candidates for each question are annotated by humans.
Overall, \gpd contains 6,939 questions and 283,855 candidate answers.
We reserve 3,000 questions for evaluation, 808 for development, and use the rest for training.
Compared to ASNQ and WikiQA, whose candidate answers are mostly from Wikipedia pages, \gpd contains answers that are from a diverse set of pages, which allow us to better estimate robustness with respect to content obtained from the web.

\section{Common Training Configurations}
Besides the method-specific hyperparameters described in Sections~\ref{subsec:distilled} and~\ref{subsec:our_model}, we describe training strategies and hyperparameters commonly used to train \astwo models in this study.
Unless we specified, we used Adam optimizer~\cite{kingma2015adam} with a linear learning rate scheduler with warm up\footnote{We used the warm up strategy only for the first 2.5 - 10\% of training iterations, using \url{https://huggingface.co/docs/transformers/main_classes/optimizer_schedules\#transformers.get_linear_schedule_with_warmup}} to train \astwo models.
The number of training iterations was 20,000, and we assess a \astwo model every 250 iterations using the dev set for validation.
If the dev MAP is not improved within the last 50 validations\footnote{For the ASNQ dataset, we considered the last 25 validations as ``patience''.}, we terminate the training session.
As described in Section~\ref{subsec:as2_results}, we independently tuned hyperparameters based on the dev set for each dataset, including an initial learning rate $\{10^{-6}, 10^{-5}\}$ and batch size $\{8, 16, 24, 32, 64\}$.
Note that we train \astwo models on the ASNQ dataset for 200,000 iterations due to the size of the dataset.

For model configurations, we used the default configurations available in Hugging Face Transformers 3.0.2~\cite{wolf2020huggingface}.
For instance, the number of attention heads are 12 and 64 for \albertB and \albertXXL, 12 and 16 for \robertaB and \robertaL, and 12 and 16 for \electraB and \electraL, respectively.
In this paper, we designed \ourmodel leveraging the default \electraB architecture, thus the number of attention heads is 12.

\end{document}